\documentclass[sigconf]{acmart}

\setcopyright{none}
\renewcommand\footnotetextcopyrightpermission[1]{}
\acmConference[ICAIF '26]{The 7th ACM International Conference on AI in Finance}{November 14--17, 2026}{Milan, Italy}
\acmYear{2026}
\acmDOI{}

\usepackage{mathtools}
\usepackage{booktabs}
\usepackage{array}
\usepackage{multirow}
\usepackage{graphicx}
\usepackage{subcaption}
\usepackage{enumitem}
\usepackage{placeins}

\graphicspath{{artifacts/}}
\microtypesetup{expansion=false}
\setlist[itemize]{leftmargin=1.35em,itemsep=2pt,topsep=3pt}
\setlist[enumerate]{leftmargin=1.55em,itemsep=2pt,topsep=3pt}
\newcommand{\qhat}{\widehat q}
\newcommand{\chat}{\widehat C}
\newcommand{\E}{\mathbb{E}}

\begin{teaserfigure}
  \includegraphics[width=\textwidth]{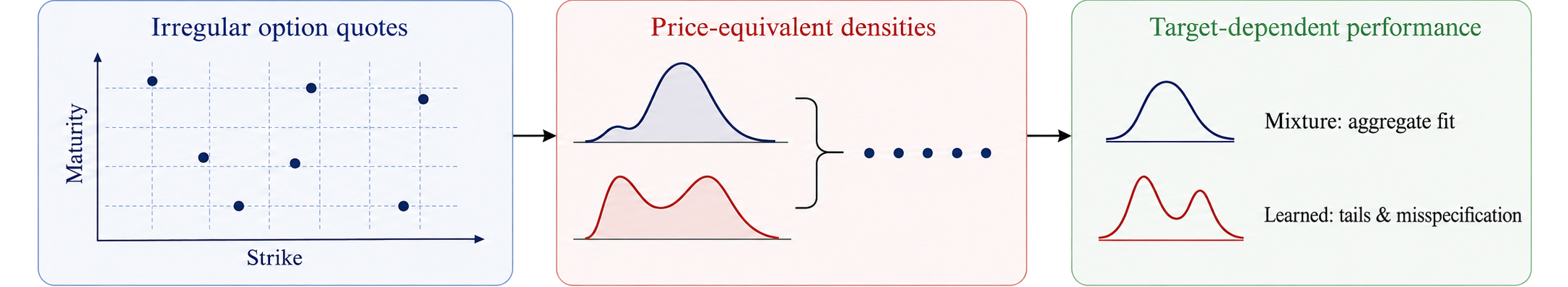}
  \caption{\textbf{Inverse learning of latent risk-neutral densities.}
  Irregular option quotes provide sparse observations across strikes and
  maturities (left), but different unobserved left tails can produce identical
  prices at the covered strikes (center). Estimator performance depends on the evaluation target (right): a well-specified lognormal mixture leads the aggregate synthetic density
  and price metrics, and learned operators improve selected tail
  functionals and recovery under structural misspecification. Accurate
  option prices therefore do not guarantee accurate recovery
  of the latent risk-neutral density.}
  \Description{A three-panel schematic connected by left-to-right arrows.
  The left blue panel, titled ``Irregular option quotes,'' shows sparse
  observations distributed across a strike-and-maturity grid. The center
  red panel, titled ``Price-equivalent densities,'' shows two visibly
  different left-tail density curves mapped to the same row of covered-strike
  option prices, illustrating extrapolative nonidentifiability. The right
  green panel, titled ``Target-dependent performance,'' contrasts a smooth
  unimodal curve labeled ``Mixture: aggregate fit'' with a multimodal curve
  labeled ``Learned: tails and misspecification,'' indicating that the
  strongest method depends on the evaluation target and data-generating
  family.}
  \label{fig:teaser}
\end{teaserfigure}

\title{Inverse Learning of Latent Risk-Neutral\\Densities from Irregular
Option Quotes}

\author{Lennon J. Shikhman}
\authornote{Corresponding author.}
\email{lj@shikhman.net}
\affiliation{
  \department{College of Computing}
  \institution{Georgia Institute of Technology}
  \country{USA}
}

\author{Michael Galarnyk}
\email{mgalarnyk3@gatech.edu}
\affiliation{
  \department{College of Engineering}
  \institution{Georgia Institute of Technology}
  \country{USA}
}

\author{Aadi Dash}
\email{aadidash@gatech.edu}
\affiliation{
  \department{College of Computing}
  \institution{Georgia Institute of Technology}
  \country{USA}
}

\author{Nicholas A. Welsh}
\email{nwelsh2024@my.fit.edu}
\affiliation{
  \department{Department of Mathematics and Systems Engineering}
  \institution{Florida Institute of Technology}
  \country{USA}
}

\begin{document}

\begin{abstract}
Accurate option prices do not imply accurate recovery of the latent
risk-neutral density.  We study this distinction with two complementary
benchmarks.  A controlled benchmark exposes simulator-truth
densities for latent evaluation, while a chronological NIFTY benchmark tests
only held-out market prices.  A two-component lognormal mixture has the lowest
aggregate price, \(L^1\), Wasserstein, and fixed-tail errors on the synthetic
benchmark.  Learned operators retain narrower strengths: DeepONet reduces
1\% quantile and variance error by \(39.0\%\) and \(34.6\%\) relative to the
mixture, and a quote transformer reduces \(L^1\) by \(16.4\%\) on the
structurally misspecified Merton family.  A numerical conditioning analysis
illustrates why these rankings can differ: after enforcing mass and forward
constraints, 95 of 126 pricing directions fall below a \(10^{-14}\) relative
threshold on the stated discretization, and two densities with different
unobserved left tails and \(L^1=0.061\) produce identical covered-strike prices.
On 524 held-out NIFTY calls, validation-selected test-time adaptation
reduces DeepONet RMSE by \(28.3\%\), but per-expiry mixture and SVI fits remain
much more accurate.  The evidence supports target-dependent inductive bias,
not a universal winner.
\end{abstract}

\keywords{risk-neutral density, option pricing, inverse problems, operator
learning, irregular observations, financial machine learning}

\ccsdesc[500]{Applied computing~Economics}
\ccsdesc[500]{Computing methodologies~Machine learning}

\maketitle

\section{Introduction}
\label{sec:introduction}

Option prices contain information about a market-implied distribution over a
future asset price.  This RND supports tail probabilities, quantiles, and
risk-neutral moments, but it is not directly observed.  Markets instead
provide prices at a changing, sparse collection of strikes and maturities.
Practical pipelines commonly smooth prices or implied volatilities and then
differentiate the fitted surface
\citep{breeden1978prices,jackwerth1996recovering,aitsahalia1998nonparametric,
tzavalis2008recovering,figlewski2008estimating,figlewski2018review,
monteiro2011estimation}.

This workflow creates two questions that are easy to conflate: whether a
method reconstructs observable prices and whether it recovers the latent
density that generated them.  The first can be tested on markets by hiding
quotes.  The second cannot because a market RND is itself inferred.  Synthetic
data are necessary for latent error, but synthetic evidence alone cannot
establish practical market accuracy.  In machine-learning terms, RND recovery
is an inverse problem from irregular observations
\citep{arridge2019inverse,putzky2019invert}, with a function-valued output that
invites operator learning \citep{lu2021deeponet,li2021fourier,
kovachki2023neural}.

We therefore separate the evidence.  The simulator draws exact densities from
martingale process families, prices continuous quotes analytically, and
reveals the density only for controlled supervision and evaluation.  The
market study uses chronologically disjoint NIFTY transaction bars and tests
only held-out prices.  Four learned backbones are compared with direct
parametric and regularized density estimators, surface smoothers, and
Breeden--Litzenberger (BL) extraction.  Matched density-output and
direct-surface variants isolate representation from architecture.

Across the controlled study, density MSE is the most effective tested
supervised objective.  A well-specified mixture is strongest on aggregate
synthetic metrics, while learned operators improve selected functionals and
the misspecified Merton regime.  Test-time adaptation helps on NIFTY but does
not close the gap to local classical refitting.  The relevant scientific
result is not a single winning architecture: the inverse problem, supervision
target, structural specification, and downstream functional jointly determine
the ranking.

This paper is a controlled empirical findings study, not a claim of a new
universal architecture.  Its contributions are:

\begin{itemize}
  \item A benchmark that separates simulator-truth latent evaluation from
  held-out real-market observable evaluation.
  \item A ten-seed density-MSE protocol with direct mixture and regularized
  density baselines, decision-relevant functionals, and family-stratified
  results.
  \item A numerical spectrum and constructive density pair that illustrate
  finite-grid conditioning and uncovered-tail ambiguity.
  \item Matched representation, stress, operator-mismatch, and real-data
  adaptation studies that delimit where the learned prior helps and fails.
\end{itemize}

\section{Related Work}
\label{sec:related}

\subsection{Risk-neutral density estimation.}
Breeden and Litzenberger identify the state-price density with a second strike
derivative \citep{breeden1978prices}.  Practical estimators combine smoothing,
interpolation, or constrained optimization with differentiation
\citep{jackwerth1996recovering,aitsahalia1998nonparametric}.  Direct recovery,
tail treatment, and maturity-indexed RND surfaces are studied in
\citep{tzavalis2008recovering,figlewski2008estimating,figlewski2018review,
monteiro2011estimation}.  Recent distributional approaches use normalizing
flows and generative networks \citep{yang2023mixture,xian2024riskneutral}.  We
study controlled recovery, not the novelty of RND estimation itself.  These
finance-specific learned estimators are not included in our benchmark: the
learned suite instead isolates backbone, output representation, and
supervision effects under a common protocol.  Consequently, our comparisons
do not establish superiority over specialized learned RND estimators.

\subsection{Surface smoothing and no-arbitrage structure.}
SVI is a standard parametric volatility-slice model with known static-arbitrage
conditions \citep{gatheral2014arbitrage}.  Roper studies continuous regularity
and arbitrage conditions \citep{roper2010arbitrage}; Ackerer et al.\ use
financial structure in neural smoothing \citep{ackerer2020deep}; and Operator
Deep Smoothing maps sparse quotes to volatility
\citep{wiedemann2025operator}.  Related neural work covers calibration,
surrogates, real-time smoothing, finance-informed losses, and term structure
\citep{hernandez2016calibration,horvath2021deep,ruf2020neural,
yang2025hyperiv,aboussalah2026finance,zhang2025riskneutral}.  Transfer learning
has been proposed for illiquid regimes \citep{conti2026transfer}.  Our discrete
certification is not equivalent to Roper's continuous conditions.

\subsection{Learning from irregular option data.}
Hexagon-Net uses heterogeneous graph attention to propagate information across
liquidity regions and forecast volatility surfaces
\citep{liang2025hexagon}, while forecasting benchmarks have used replayable
historical evaluation with evidence restricted to information available at each
forecast date \citep{ye2026forecastdojoreplayableenvironmentsbenchmarking}.
These forecasting targets differ from our same-snapshot inverse target, but
they motivate careful treatment of irregular quote geometry and evaluation
design.

\subsection{Operator and set architectures.}
DeepONet couples branch and trunk networks \citep{lu2021deeponet}; FNO uses
spectral convolutions \citep{li2021fourier}; and related operator work studies
latent representations, general geometries, convolutional constructions, boundary-indexed operators, and
pretraining \citep{kovachki2023neural,wang2024latent,shikhman2026one,raonic2023convolutional,chen2024dataefficient}.  Deep Sets and Transformers
provide invariant or attention-based encoders
\citep{zaheer2017deep,vaswani2017attention}.  Learned inversion and
physics-informed models motivate differentiating through known forward maps
\citep{putzky2019invert,raissi2019physics}, while structured-shift studies
warn that robustness depends on both architecture and problem family
\citep{shikhman2026diagnosing}.  We use these as controlled backbones rather
than architectural contributions.

\section{Problem and Evaluation Principle}
\label{sec:problem}

\subsection{Normalized option-pricing map}

For maturity \(\tau\), let \(F_\tau\) be the forward price,
\(x=\log(S_\tau/F_\tau)\) normalized log terminal price, and
\(k=\log(K/F_\tau)\) log-moneyness.  A normalized risk-neutral density
\(q_\tau(x)\) satisfies
\begin{equation}
q_\tau(x)\ge 0,\qquad
\int q_\tau(x)\,dx=1,\qquad
\int e^x q_\tau(x)\,dx=1.
\label{eq:density_constraints}
\end{equation}
With discounting removed by forward normalization, the call price is
\begin{equation}
C(k,\tau)
=\int (e^x-e^k)_+q_\tau(x)\,dx.
\label{eq:pricing}
\end{equation}
The observed input is an unordered set
\begin{equation}
\mathcal Q
=\{(k_i,\tau_i,\widetilde C_i)\}_{i=1}^{n},
\qquad
\widetilde C_i=C(k_i,\tau_i)+\epsilon_i,
\label{eq:quotes}
\end{equation}
where both \(n\) and the locations vary by example.

\subsection{Two outputs and two kinds of evidence}

A density-output model predicts \(\qhat_\tau(x)\); the differentiable version
of Eq.~\eqref{eq:pricing} maps it to \(\chat(k,\tau)\).  A direct-surface model
predicts \(\chat(k,\tau)\) first.  For a controlled latent comparison, we apply
the same fixed physical-strike BL extraction, clipping, trapezoidal
normalization, and forward-moment repair to every direct surface.

Simulator-truth density error is available only in the synthetic study.
Market data support held-out price and implied-volatility tests, not a claim
that an inferred density equals a unique ``true'' market density.  We enforce
this boundary throughout the results.

\subsection{Metrics}
\label{sec:metrics}

Let \(Q_\tau(x)=\int_{-\infty}^{x}q_\tau(u)\,du\).  We report maturity-averaged
integrated density error
\begin{equation}
L^1_q
=\frac{1}{|\mathcal T|}
\sum_{\tau\in\mathcal T}\int
|\qhat_\tau(x)-q_\tau(x)|\,dx,
\end{equation}
and the one-dimensional Wasserstein distance
\begin{equation}
W_1
=\frac{1}{|\mathcal T|}
\sum_{\tau\in\mathcal T}\int
|\widehat Q_\tau(x)-Q_\tau(x)|\,dx.
\label{eq:w1}
\end{equation}
Equation~\eqref{eq:w1} is the cumulative-distribution term: it evaluates mass
allocation across the domain rather than duplicating pointwise density MSE
\citep{peyre2019computational}.  The tail metric is the absolute error in
\(\Pr[x<-0.5]\), averaged by maturity.  We add absolute error in the 1\%
quantile and in risk-neutral variance as decision-relevant latent functionals.
Observable metrics are price RMSE and Black--Scholes implied-volatility RMSE
on valid inversions \citep{black1973pricing}.  Synthetic price error uses a
dense common grid; real price error uses quotes withheld before fitting.

\subsection{Numerical conditioning of the pricing map}
\label{sec:conditioning}

To connect the evaluation principle to the inverse mechanism, we discretize
Eq.~\eqref{eq:pricing} on the 128-node density and strike grids.  We restrict
perturbations to the 126-dimensional subspace with zero total mass and zero
forward moment, then compute its singular spectrum.  At a relative threshold
of \(10^{-14}\), only 31 directions are resolved and 95 fall below threshold;
the condition number within the resolved subspace is \(6.89\times10^3\).
This count is specific to the density and strike grids, covered strike range,
quadrature, normalization, weighting, and threshold; it is not an
identification theorem.  Uniform scaling leaves the relative spectrum
unchanged, whereas nonuniform weighting or discretization need not.
Observation noise does not change these singular values but does change which
directions are practically resolvable.  Because the map is analyzed
maturity-wise, additional maturities help only through imposed cross-maturity
structure.

\begin{figure}[h]
  \centering
  \includegraphics[width=\linewidth]{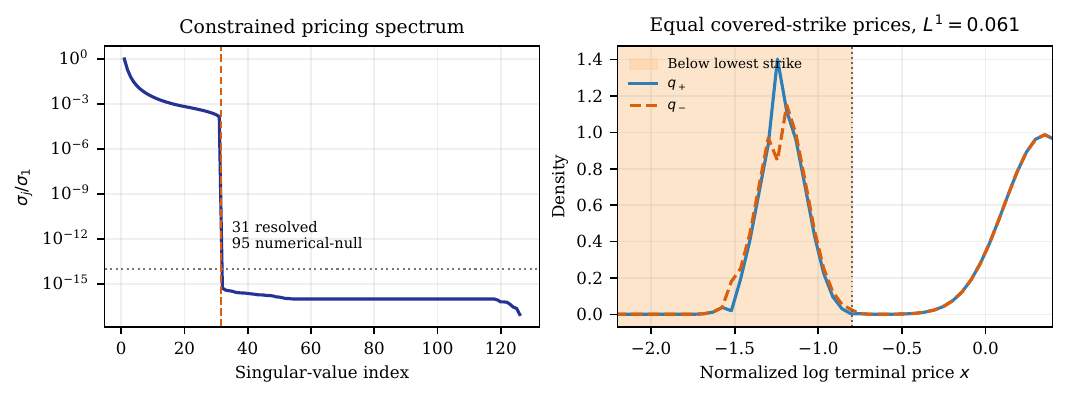}
  \Description{A log-scale constrained singular-value spectrum beside two
  distinct risk-neutral densities whose option prices coincide on the observed
  strike range.}
  \caption{Left: constrained singular spectrum for the stated discretization
  and relative threshold. Right: two nonnegative, mass- and forward-matched
  densities that differ below the lowest observed strike. Their integrated
  \(L^1\) distance is 0.061, while their prices agree to machine precision at
  every covered strike. The pair demonstrates uncovered-left-tail
  nonidentifiability, not ambiguity confined to the covered region.}
  \label{fig:conditioning}
\end{figure}

\section{Methods and Experimental Design}
\label{sec:methods}

\subsection{Synthetic benchmark}
\label{sec:synthetic_design}

\paragraph{Simulator truth.}
Each synthetic example is drawn directly from one of three positive
martingale process families: time-varying-volatility geometric Brownian
motion, an inception-time mixture of lognormal processes, or compensated
Merton jump diffusion \citep{black1973pricing,merton1976option}.  The density
is the generating distribution, not the output of a smoothing method.
Headline quote prices are evaluated with family-specific analytic formulas.
The learning target is represented on 128 \(x\)-nodes over \([-4,3]\), while
an independent 1,024-node audit grid covers \([-8,5]\).  Across the three
families, analytic-versus-audit-grid price RMSE is approximately
\(9\!\times\!10^{-6}\), and analytic-versus-model-grid RMSE ranges from
\(1.15\!\times\!10^{-4}\) to \(1.27\!\times\!10^{-4}\).  Thus the quotes are
not generated by the 128-node pricing layer used during learning.

\paragraph{Sampling and noise.}
Every seed generates 3,000 examples with a 70/15/15 split
(2,100/450/450).  Each surface has 16 maturities and a 128-node strike grid,
but only 40--320 continuous, jittered strike--maturity locations are observed,
corresponding to 2--15.625\% of the 2,048-node lattice.  Across the 4,500 test
examples, the realized mean is 180.24 quotes.  Six regimes are mixed:
uniform, maturity-stratified, strike-local, tail-undersampled, clustered, and
market-like.  ``Market-like'' names a synthetic stress geometry only: it is
neither estimated from nor calibrated to NIFTY or exchange sampling
frequencies.  Bounded heteroskedastic noise depends on
moneyness, maturity, and liquidity proxies and is sampled conditional on
static call bounds; quotes are not clipped after corruption.

\subsection{Learned models and classical baselines}
\label{sec:models}

The four backbones share a padded quote-set interface.  DeepONet combines a
mean-pooled quote branch with a coordinate trunk.  FNO averages quote
embeddings onto the output grid and applies four two-dimensional spectral
blocks.  The quote transformer uses two layers and four attention heads.  The
set decoder uses a pointwise MLP, invariant mean pooling, and coordinate
decoder.

The primary benchmark trains density outputs for all four.  Its strongest
classical comparator is a joint two-component martingale-lognormal density
fitted to all irregular quotes by soft-\(L^1\) least squares with six fixed
starts.  It nests the Black--Scholes family and matches the simulator's
two-component family, but is structurally misspecified for Merton jumps.  A
second direct baseline fits a nonnegative discrete density at each observed
maturity using a second-difference penalty plus mass and forward penalties,
then interpolates across maturities.  The real-data comparison additionally
includes interpolation, SVI, a shape-constrained spline, and BL extraction.

\subsection{Objective and financial structure}
\label{sec:losses}

The primary protocol uses only supervised density MSE,
\begin{equation}
\mathcal L_{\mathrm{main}}=\mathcal L_{\mathrm{pdf}}.
\label{eq:loss}
\end{equation}
An objective ablation compares density MSE with combinations of observed-quote
MSE, Eq.~\eqref{eq:w1}, a forward-moment penalty, and an \(H^1\)-style grid
roughness penalty.  Density MSE provides the strongest latent recovery under
the tested configuration.  The forward term is
\begin{equation}
\mathcal L_{\mathrm{fwd}}
=\E_{\tau}\left[
\left(\int e^x\qhat_\tau(x)\,dx-1\right)^2
\right].
\end{equation}
It enforces one martingale moment, not complete no-arbitrage.  Likewise, the
roughness term regularizes a finite density grid and does not establish twice
differentiability of implied volatility.

For reporting, predicted prices are certified against bounds, strike
monotonicity, vertical-spread slopes, physical-strike convexity, and calendar
monotonicity.  A deterministic PAVA-based repair is reproducible but is not
the exact global Euclidean projection; an optional quadratic program supplies
that projection.  These checks relate to arbitrage detection and learned
projection
\citep{cohen2020detecting,yang2020projections}.  They complement, but do not
replace, the continuous conditions studied by \citet{roper2010arbitrage} and
\citet{ackerer2020deep}.

\subsection{Controlled comparisons and stress tests}

The price-only representation comparison trains density and direct-surface
variants with the same observed-price loss, data, seeds, minibatch order,
initial weights, budget, and parameter count.  A separately labeled oracle
comparison aligns dense targets to each representation.  An auxiliary
robustness study evaluates 31 stress settings and three family holdouts for
seeds 42--51, for 34 conditions in total.  It uses the composite objective and BL-style classical
baselines, so it is interpreted as a sensitivity analysis rather than as the
primary comparison against the direct mixture estimator.

\subsection{Real NIFTY held-out evaluation}
\label{sec:nifty_method}

We use the CC0 NIFTY spot, futures, and options one-minute archive of
\citet{bhat2024nifty}.  The adapter selects positive-volume 15:15
Asia/Kolkata transaction-bar closes from 2019--2020, calls with 7--35 days to
maturity, and an exact-minute maturity-aligned NIFTY F1 close.  It normalizes
by \(F=\mathrm{F1}\) and uses \(D=1\), an explicit short-horizon approximation.
These are transaction bars, not bid--ask quotes, NBBO midpoints, or executable
prices.  Exact-minute closes need not occur at the same second.  Because every
method uses the same normalization, this approximation primarily limits the
absolute financial interpretation of the benchmark; it need not preserve
rankings under a different market-data construction.

The processed sample contains 6,191 calls, 295 dates, and 19 expiries, with
295 post-hoc call--put parity audits.  Splits are chronological:
2,667 calls on 165 dates for training through 19 December 2019; 1,947 calls on
66 dates for validation through 23 April 2020; and 1,577 calls on 64 dates
from 4 May through 20 August 2020.  Within every test chain, a fixed
moneyness-stratified mask reveals 1,053 prices and withholds 524
(\(33\%\)).  No latent labels are created.

All eight learned variants are trained only on observed real prices for 50
epochs and ten initialization seeds.
Classical comparators include strike interpolation, two-dimensional surface
interpolation, SVI, a shape-constrained spline, BL constrained-spline density,
a two-component lognormal mixture, and a regularized discrete density.  The
classical methods are fitted independently to each observed expiry slice;
the learned models amortize across the training period.

To make this comparison more structurally informative, we adapt the pretrained
DeepONet surface separately to each observed test chain and evaluate only its
hidden quotes.  Five step-size and iteration candidates are selected once on
the chronological validation period.  The selected setting uses 50 Adam steps,
learning rate \(10^{-4}\), and a \(10^{-4}\) proximal penalty to the pretrained
weights.  Each test surface starts from the same seed-specific checkpoint, so
adaptation cannot carry information across test dates.

\begin{figure*}[h]
  \centering
  \includegraphics[width=\textwidth]{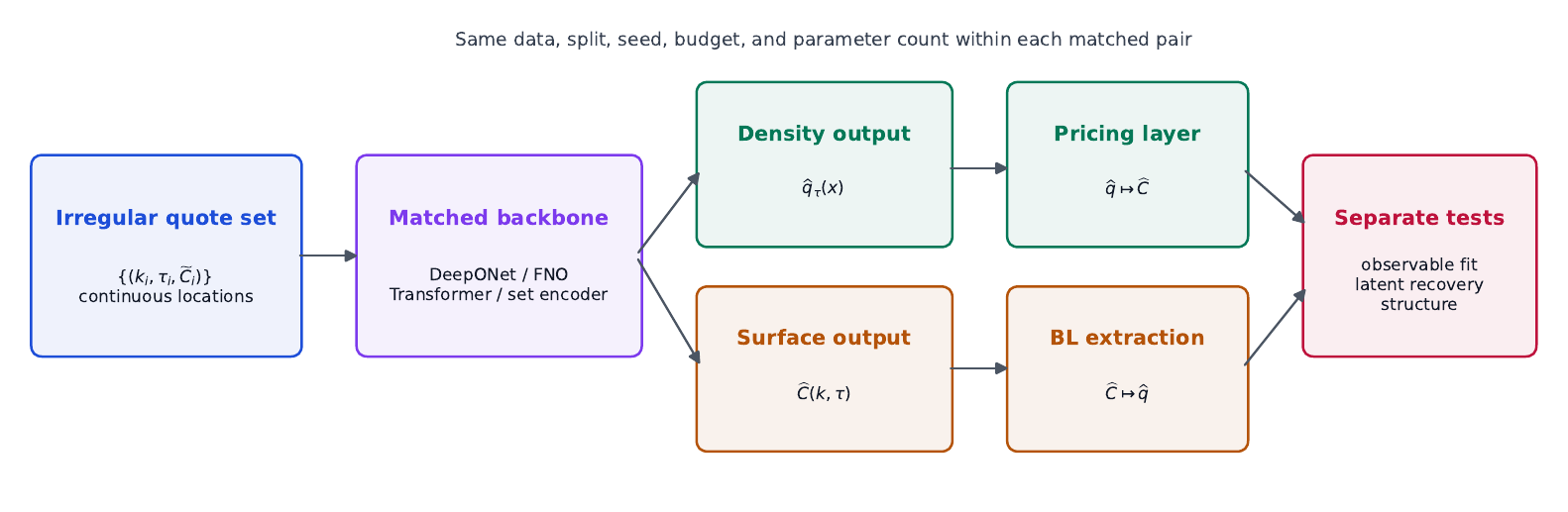}
  \Description{A flow diagram separating synthetic latent-density training and
  evaluation from held-out real-market option-price evaluation.}
  \caption{Evaluation pipeline.  Simulator densities are available only to
  synthetic training and latent evaluation.  Real-market evaluation withholds
  observable quotes and creates no latent labels.}
  \label{fig:pipeline}
\end{figure*}

\subsection{Uncertainty and reproducibility}
\label{sec:statistics}

Synthetic tables use one rerun seed as the unit and report the mean with a
95\% nonparametric bootstrap interval over ten seeds
\citep{efron1993bootstrap}.  We prespecified three primary paired comparisons:
set decoder versus mixture lognormal for \(L^1\), \(W_1\), and fixed-tail
error.  Exact sign-flip tests are Holm-adjusted within this three-test family
\citep{holm1979simple}.  Quantile, variance, regularized-density, and
family-stratified comparisons are labeled exploratory.  With ten pairs, the
smallest two-sided exact \(p\)-value is \(0.001953\).

For real learned models, intervals vary the ten initialization seeds.  The
classical baselines are deterministic under the frozen mask, so repeating
their value across seeds would create false precision; their intervals instead
cluster-bootstrap the 64 calendar dates.  Adapted versus zero-shot DeepONet is
paired by seed.  The two uncertainty units are reported separately.

\section{Results}
\label{sec:results}

\subsection{Objective selection and stronger density baselines}
\label{sec:main_results}

\begin{table*}[h]
\centering
\scriptsize
\setlength{\tabcolsep}{2.5pt}
\caption{Main synthetic benchmark with density-MSE supervision (3,000 surfaces per seed; ten seeds). Entries are seed means with 95\% bootstrap intervals. Boldface marks the minimum in each metric column; no method minimizes every decision-relevant error.}
\label{tab:main_results}
\resizebox{\textwidth}{!}{%
\begin{tabular}{lcccccc}
\toprule
Method & Price RMSE $\downarrow$ & Density $L^1$ $\downarrow$ & $W_1$ $\downarrow$ & Tail $\downarrow$ & $1\%$ quantile $\downarrow$ & Variance $\downarrow$ \\
\midrule
DeepONet & 0.0087 {\scriptsize [0.0073, 0.0103]} & 0.1620 {\scriptsize [0.1575, 0.1666]} & 0.0278 {\scriptsize [0.0267, 0.0290]} & 0.0153 {\scriptsize [0.0147, 0.0160]} & \textbf{0.1348 {\scriptsize [0.1294, 0.1405]}} & \textbf{0.0137 {\scriptsize [0.0130, 0.0148]}} \\
FNO & 0.0205 {\scriptsize [0.0077, 0.0393]} & 0.1564 {\scriptsize [0.1275, 0.1971]} & 0.0347 {\scriptsize [0.0221, 0.0532]} & 0.0165 {\scriptsize [0.0115, 0.0236]} & 0.3021 {\scriptsize [0.1027, 0.5938]} & 0.0366 {\scriptsize [0.0124, 0.0720]} \\
Quote transformer & 0.0108 {\scriptsize [0.0098, 0.0119]} & 0.1374 {\scriptsize [0.1337, 0.1413]} & 0.0257 {\scriptsize [0.0250, 0.0266]} & 0.0164 {\scriptsize [0.0152, 0.0177]} & 0.1547 {\scriptsize [0.1466, 0.1624]} & 0.0192 {\scriptsize [0.0182, 0.0202]} \\
Set decoder & 0.0158 {\scriptsize [0.0145, 0.0173]} & 0.1683 {\scriptsize [0.1642, 0.1729]} & 0.0305 {\scriptsize [0.0295, 0.0315]} & 0.0158 {\scriptsize [0.0148, 0.0168]} & 0.1512 {\scriptsize [0.1449, 0.1573]} & 0.0222 {\scriptsize [0.0206, 0.0238]} \\
\midrule
Mixture lognormal & \textbf{0.0033 {\scriptsize [0.0032, 0.0033]}} & \textbf{0.1262 {\scriptsize [0.1235, 0.1290]}} & \textbf{0.0245 {\scriptsize [0.0240, 0.0250]}} & \textbf{0.0116 {\scriptsize [0.0113, 0.0120]}} & 0.2211 {\scriptsize [0.2146, 0.2275]} & 0.0210 {\scriptsize [0.0207, 0.0214]} \\
Regularized discrete & 0.0096 {\scriptsize [0.0095, 0.0098]} & 0.2775 {\scriptsize [0.2752, 0.2797]} & 0.0630 {\scriptsize [0.0623, 0.0637]} & 0.0199 {\scriptsize [0.0195, 0.0202]} & 1.1525 {\scriptsize [1.1270, 1.1813]} & 0.1240 {\scriptsize [0.1222, 0.1256]} \\
\bottomrule
\end{tabular}
}
\end{table*}

An objective comparison shows that density MSE reduces \(L^1\) relative to the
composite objective by 57.3\% for DeepONet, 58.4\% for FNO, 63.7\% for the
quote transformer, and 55.0\% for the set decoder.  Every reduction occurs in
all ten paired seeds.  Under density-MSE supervision, the quote transformer
has the lowest stable learned \(L^1\) and \(W_1\), while DeepONet has the best
learned fixed-tail, 1\% quantile, variance, and price errors.  FNO has two
failure seeds and correspondingly wide intervals.

The direct mixture baseline is stronger on aggregate price, \(L^1\), \(W_1\),
and fixed-tail error.  All three prespecified confirmatory comparisons favor
the mixture: set-decoder error is 33.4\%, 24.5\%, and 36.0\% higher,
respectively, with exact Holm-adjusted \(p=0.0059\) for each.  Thus, the learned
models are not the strongest general density estimators under the aggregate
metrics.

\begin{figure}[h]
  \centering
  \includegraphics[width=\linewidth]{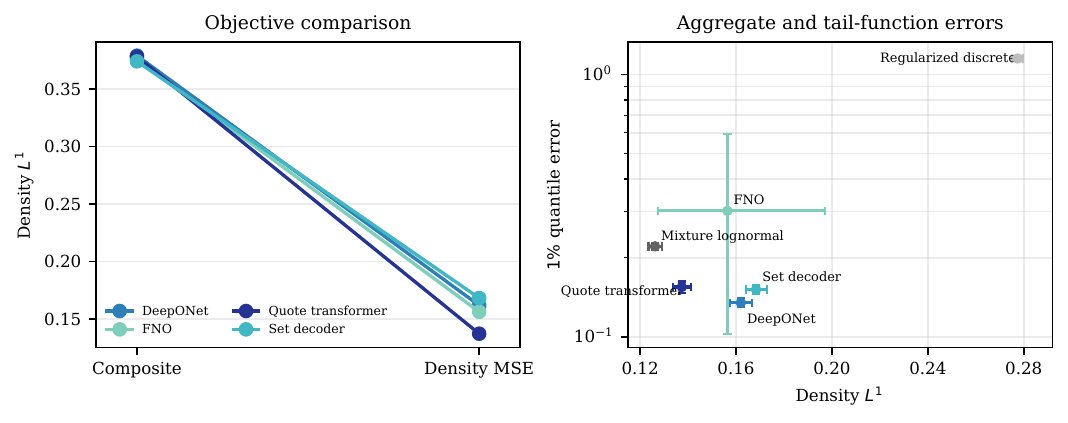}
  \Description{Paired comparisons of density objectives and a scatter plot
  showing the tradeoff between aggregate density error and one-percent
  quantile error.}
  \caption{Left: changing only the training objective lowers \(L^1\) for all
  four learned backbones. Right: aggregate \(L^1\) and 1\% quantile error
  select different methods; points show seed means and 95\% intervals.}
  \label{fig:tradeoff}
\end{figure}

\subsection{Where the learned prior helps}
\label{sec:robustness_results}

Aggregate performance conceals economically relevant differences.  DeepONet's
1\% quantile error is 39.0\% below the mixture and its variance error is
34.6\% lower; both exploratory paired comparisons favor DeepONet in all ten
seeds (unadjusted exact \(p=0.001953\)).  Family stratification identifies the
source.  The mixture is strongest on Black--Scholes and mixture draws, where
its form is well matched.  On Merton jump diffusion, the quote transformer
reduces \(L^1\) by 16.4\% relative to the mixture, while DeepONet improves
\(W_1\) and fixed-tail error.  These are specification-sensitive gains, not
evidence of universal learned superiority.

The learned models also beat the regularized discrete estimator: the set
decoder reduces \(L^1\), \(W_1\), and tail error by 39.3\%, 51.6\%, and
20.4\%.  An auxiliary 34-condition stress suite uses the composite objective
and omits the direct mixture baseline, so we interpret it only as sensitivity
analysis.

\subsection{Matched representations: density output is not sufficient}
\label{sec:representation_results}

The matched price-only comparison remains mixed.  FNO favors density output;
DeepONet and the quote transformer favor direct surfaces followed by fixed BL
extraction; and the set decoder splits by metric.  Under representation-aligned
dense oracle labels, density outputs improve latent metrics for all four
backbones.  Thus representation, deterministic output map, and supervision
interact; density output alone is not sufficient.

\subsection{Additional controls}
\label{sec:ablation_results}

A 1,000-surface objective ablation and the 3,000-surface objective comparison
both show that the tested auxiliary losses hurt \(L^1\).  Price-only density
training remains much worse.  Increasing the set-decoder sample size from
1,000 to 3,000 improves \(L^1\), with a smaller gain at 5,000.  Separately,
replacing the 128-node training pricer with analytic prices changes DeepONet
\(L^1\) by only \(3\times10^{-5}\).  This controls numerical resolution, not
stochastic-model misspecification.

\subsection{Real NIFTY data: adaptation helps but does not close the gap}
\label{sec:real_results}

\begin{table*}[h]
\centering
\scriptsize
\setlength{\tabcolsep}{3.0pt}
\caption{NIFTY held-out transaction-price results (524 hidden calls over 64 dates). Learned intervals vary initialization seeds; deterministic classical intervals resample dates, so the two interval types are not interchangeable.}
\label{tab:real_results}
\begin{tabular}{llcc}
\toprule
Method & Regime & Price RMSE $\downarrow$ & Price MAE $\downarrow$ \\
\midrule
Mixture lognormal & Classical refit & 0.000166 {\scriptsize [0.000142, 0.000195]} & 0.000107 {\scriptsize [0.000097, 0.000118]} \\
SVI smile & Classical refit & 0.000241 {\scriptsize [0.000190, 0.000291]} & 0.000145 {\scriptsize [0.000119, 0.000173]} \\
Shape-constrained spline & Classical refit & 0.001086 {\scriptsize [0.000901, 0.001289]} & 0.000656 {\scriptsize [0.000597, 0.000717]} \\
BL constrained density & Classical refit & 0.001712 {\scriptsize [0.001480, 0.001940]} & 0.001219 {\scriptsize [0.001064, 0.001389]} \\
DeepONet surface + adaptation & Learned + refit & 0.003464 {\scriptsize [0.003207, 0.003758]} & 0.002594 {\scriptsize [0.002468, 0.002720]} \\
DeepONet surface & Learned zero-shot & 0.004833 {\scriptsize [0.004149, 0.005615]} & 0.003557 {\scriptsize [0.003025, 0.004157]} \\
Regularized discrete density & Classical refit & 0.005464 {\scriptsize [0.004831, 0.006069]} & 0.004379 {\scriptsize [0.003824, 0.004949]} \\
Strike-linear interpolation & Classical & 0.010725 {\scriptsize [0.006068, 0.014726]} & 0.001925 {\scriptsize [0.001198, 0.002758]} \\
\bottomrule
\end{tabular}
\end{table*}

Validation-selected adaptation lowers DeepONet surface RMSE from 0.004833 to
0.003464, a 28.3\% paired reduction that occurs for all ten seeds (exact
\(p=0.001953\)).  MAE falls by 27.1\% in eight seeds
(\(p=0.0078\)).  Adaptation takes 0.138 seconds per surface on average and now
beats the regularized density and simple interpolation in RMSE.  The change in
price-bound violations is inconclusive.  Because comparator runtimes were not
benchmarked under a common implementation, the adaptation timing does not by
itself establish an end-to-end computational advantage.

The strongest local fits remain far better: mixture lognormal and SVI attain
RMSE 0.000166 and 0.000241.  The adapted network is also worse than the
shape-constrained spline and BL constrained density.  Classical models refit
each expiry, whereas the neural model uses amortized pretraining plus a short
refit; this is a more informative comparison, but not identical statistical
machinery.  These results validate observable prices only, not a unique
market-truth RND.

\section{Discussion}
\label{sec:discussion}

The conditioning experiment provides a finite-grid illustration of the
rankings.  Pricing integrates a density against call payoffs, while BL
extraction differentiates a fitted surface twice.  In the present construction,
the equal-price pair differs below the lowest covered strike and therefore
demonstrates extrapolative left-tail nonidentifiability; it does not establish
comparable ambiguity entirely within the covered region.  More generally,
weakly observed finite-grid directions require an estimator to select among
price-compatible densities through an explicit or learned structural prior.
The mixture prior is excellent when its family is well matched.  The learned
prior becomes useful for the Merton family and for functionals such as extreme
quantiles and variance.  Observable fit alone cannot identify which prior
recovers the generating density.

Simulation supplies
latent labels for controlled measurement. At inference every trained model
receives only quotes.  Market data supply a separate test of observable
generalization.  Adaptation shows that amortized training can provide a useful
initialization, but the NIFTY result still favors per-expiry classical fits.
Together with the mixed representation comparison, this rules out a claim
that density output or operator learning is intrinsically superior.  We do not
test whether simulator-density pretraining transfers across the simulator--market shift; combining such pretraining with price-only market adaptation is
a natural next experiment.

\subsection{Limitations}
\label{sec:limitations}

\begin{itemize}
  \item Simulator truth is exact only conditional on the chosen families.
  Numerical pricer mismatch is small, but economic misspecification in rates,
  dividends, exercise style, dynamics, and observation noise remains.
  \item The market study covers one index, 19 expiries, and one period.  NIFTY
  observations are transaction-bar closes without bid and ask, and exact-minute
  futures and option closes need not be synchronized to the second.  The shared
  \(D=1\) approximation, F1 normalization, time filter, and liquidity filters
  limit absolute interpretation; no market latent labels exist.
  \item Classical methods refit each expiry.  Adaptation narrows but does not
  eliminate this regime difference, and the reported adaptation time is not a
  common-system runtime comparison.  Larger multi-market training sets, time
  context, or graph encoders such as Hexagon-Net may alter the tradeoff.
  \item Ten seeds limit exact-test resolution.  The auxiliary stress suite uses
  a different objective and baseline set from the primary benchmark.
  Certification is finite-grid evidence, not a proof of continuous
  no-arbitrage.
  \item The generic learned suite does not benchmark specialized learned RND
  or volatility-smoothing estimators.  Broader market validation,
  simulator-to-market transfer, and uncertainty-aware densities or prediction
  sets remain future work.
\end{itemize}

\subsection{Broader impact.}
An RND estimator can support market monitoring, scenario analysis, and
research on implied tail risk.  The main risk is false precision: a plausible
density can be overinterpreted as a uniquely identified forecast or physical
probability.  We mitigate this by separating market observables from latent
simulator truth, reporting failures as well as wins, and avoiding trading or
decision recommendations.

\section{Conclusion}

Recovering a risk-neutral density from sparse option quotes is not equivalent
to reconstructing the observed price surface.  Our conditioning analysis makes
this distinction explicit for the stated finite grid: after enforcing mass and
forward constraints, many directions fall below the selected relative
threshold, and different unobserved left tails can produce identical prices at
the covered strikes.  This construction establishes extrapolative tail
nonidentifiability, not ambiguity confined to the covered region.  Every
estimator must therefore resolve the remaining inverse ambiguity through an
implicit or explicit structural prior.  The appropriate prior depends on the
target.  A well-specified lognormal mixture provides the strongest
aggregate synthetic performance, whereas learned operators offer narrower
advantages for tail functionals and under Merton-family misspecification.  The
matched representation experiments further show that density output alone is
not sufficient; representation, supervision, architecture, and postprocessing
jointly determine recovery quality.

The chronological NIFTY study reinforces the separation between latent and
observable evaluation.  Test-time adaptation substantially improves the
amortized DeepONet, but per-expiry mixture and SVI fits remain more accurate on
held-out market prices.  Thus, the evidence supports neither a universal
learned estimator nor a universal classical one.  RND recovery claims should
instead state the supervision available, the assumed distributional structure,
the downstream functional of interest, and whether validation concerns
observable prices or simulator-truth densities.  More broadly, price accuracy
should be treated as necessary evidence of market consistency, not as
sufficient evidence that the latent risk-neutral distribution has been
identified.

\begin{acks}
\noindent \textbf{Funding.} Lennon J. Shikhman and Michael Galarnyk acknowledge computational support from the Dell Pro Precision divison of Dell Technologies.

\noindent \textbf{Computational resources.} The authors acknowledge Dell Technologies for providing a Dell
Pro Max T2 workstation equipped with an Intel Core Ultra 9 285K
processor, 128\,GB of DDR5 ECC memory, and an NVIDIA RTX PRO
6000 Blackwell GPU, and a Dell Pro Max 16 Plus laptop equipped
with an Intel Core Ultra 9 285HX processor, 128\,GB of DDR5
memory, and an NVIDIA RTX PRO 5000 Blackwell GPU, in support
of this research.
\end{acks}

\newpage
\bibliographystyle{ACM-Reference-Format}
\bibliography{references}

@article{black1973pricing,
  author  = {Black, Fischer and Scholes, Myron},
  title   = {The Pricing of Options and Corporate Liabilities},
  journal = {Journal of Political Economy},
  volume  = {81},
  number  = {3},
  pages   = {637--654},
  year    = {1973},
  doi     = {10.1086/260062}
}

@inproceedings{shikhman2026one,
    title={One Operator to Rule Them All? On Boundary-Indexed Operator Families in Neural {PDE} Solvers},
    author={Lennon Shikhman},
    booktitle={AI{\&}PDE: ICLR 2026 Workshop on AI and Partial Differential Equations},
    year={2026},
    url={https://openreview.net/forum?id=lDjWQ9UxRy}
}

@article{breeden1978prices,
  author  = {Breeden, Douglas T. and Litzenberger, Robert H.},
  title   = {Prices of State-Contingent Claims Implicit in Option Prices},
  journal = {The Journal of Business},
  volume  = {51},
  number  = {4},
  pages   = {621--651},
  year    = {1978},
  doi     = {10.1086/296025}
}

@article{jackwerth1996recovering,
  author  = {Jackwerth, Jens Carsten and Rubinstein, Mark},
  title   = {Recovering Probability Distributions from Option Prices},
  journal = {The Journal of Finance},
  volume  = {51},
  number  = {5},
  pages   = {1611--1631},
  year    = {1996},
  doi     = {10.1111/j.1540-6261.1996.tb05219.x}
}

@article{aitsahalia1998nonparametric,
  author  = {A{\"i}t-Sahalia, Yacine and Lo, Andrew W.},
  title   = {Nonparametric Estimation of State-Price Densities Implicit in Financial Asset Prices},
  journal = {The Journal of Finance},
  volume  = {53},
  number  = {2},
  pages   = {499--547},
  year    = {1998},
  doi     = {10.1111/0022-1082.215228}
}

@article{monteiro2011estimation,
  author  = {Monteiro, Ana Margarida and T{\"u}t{\"u}nc{\"u}, Reha H. and Vicente, Lu{\'i}s Nunes},
  title   = {Estimation of Risk-Neutral Density Surfaces},
  journal = {Computational Management Science},
  volume  = {8},
  number  = {4},
  pages   = {387--414},
  year    = {2011},
  doi     = {10.1007/s10287-010-0126-3}
}

@article{merton1976option,
  author  = {Merton, Robert C.},
  title   = {Option Pricing When Underlying Stock Returns Are Discontinuous},
  journal = {Journal of Financial Economics},
  volume  = {3},
  number  = {1--2},
  pages   = {125--144},
  year    = {1976},
  doi     = {10.1016/0304-405X(76)90022-2}
}

@unpublished{roper2010arbitrage,
  author = {Roper, Michael},
  title  = {Arbitrage-Free Implied Volatility Surfaces},
  year   = {2010},
  note   = {Preprint, School of Mathematics and Statistics, The University of Sydney},
  url    = {https://www.maths.usyd.edu.au/u/pubs/publist/preprints/2010/roper-9.pdf}
}

@article{gatheral2014arbitrage,
  author  = {Gatheral, Jim and Jacquier, Antoine},
  title   = {Arbitrage-Free {SVI} Volatility Surfaces},
  journal = {Quantitative Finance},
  volume  = {14},
  number  = {1},
  pages   = {59--71},
  year    = {2014},
  doi     = {10.1080/14697688.2013.819986}
}

@inproceedings{ackerer2020deep,
  author    = {Ackerer, Damien and Tagasovska, Natasa and Vatter, Thibault},
  title     = {Deep Smoothing of the Implied Volatility Surface},
  booktitle = {Advances in Neural Information Processing Systems},
  volume    = {33},
  pages     = {11552--11563},
  year      = {2020},
  url       = {https://proceedings.neurips.cc/paper_files/paper/2020/file/858e47701162578e5e627cd93ab0938a-Paper.pdf}
}

@inproceedings{wiedemann2025operator,
  author    = {Wiedemann, Ruben and Jacquier, Antoine and Gonon, Lukas},
  title     = {Operator Deep Smoothing for Implied Volatility},
  booktitle = {International Conference on Learning Representations},
  year      = {2025},
  url       = {https://openreview.net/forum?id=DPlUWG4WMw}
}

@inproceedings{liang2025hexagon,
  author    = {Liang, Kaiwei and Liu, Ruirui and Huang, Huichou and Ruf, Johannes and Zhao, Peilin and Wu, Qingyao},
  title     = {Hexagon-Net: Heterogeneous Cross-View Aligned Graph Attention Networks for Implied Volatility Surface Prediction},
  booktitle = {Proceedings of the 31st ACM SIGKDD Conference on Knowledge Discovery and Data Mining},
  pages     = {1671--1682},
  year      = {2025},
  doi       = {10.1145/3711896.3736996}
}

@article{lu2021deeponet,
  author  = {Lu, Lu and Jin, Pengzhan and Pang, Guofei and Zhang, Zhongqiang and Karniadakis, George Em},
  title   = {Learning Nonlinear Operators via {DeepONet} Based on the Universal Approximation Theorem of Operators},
  journal = {Nature Machine Intelligence},
  volume  = {3},
  number  = {3},
  pages   = {218--229},
  year    = {2021},
  doi     = {10.1038/s42256-021-00302-5}
}

@inproceedings{li2021fourier,
  author    = {Li, Zongyi and Kovachki, Nikola and Azizzadenesheli, Kamyar and Liu, Burigede and Bhattacharya, Kaushik and Stuart, Andrew and Anandkumar, Anima},
  title     = {Fourier Neural Operator for Parametric Partial Differential Equations},
  booktitle = {International Conference on Learning Representations},
  year      = {2021},
  url       = {https://openreview.net/forum?id=c8P9NQVtmnO}
}

@inproceedings{zaheer2017deep,
  author    = {Zaheer, Manzil and Kottur, Satwik and Ravanbakhsh, Siamak and P{\'o}czos, Barnab{\'a}s and Salakhutdinov, Ruslan and Smola, Alexander J.},
  title     = {Deep Sets},
  booktitle = {Advances in Neural Information Processing Systems},
  volume    = {30},
  year      = {2017},
  url       = {https://papers.nips.cc/paper_files/paper/2017/hash/f22e4747da1aa27e363d86d40ff442fe-Abstract.html}
}

@inproceedings{vaswani2017attention,
  author    = {Vaswani, Ashish and Shazeer, Noam and Parmar, Niki and Uszkoreit, Jakob and Jones, Llion and Gomez, Aidan N. and Kaiser, {\L}ukasz and Polosukhin, Illia},
  title     = {Attention Is All You Need},
  booktitle = {Advances in Neural Information Processing Systems},
  volume    = {30},
  year      = {2017},
  url       = {https://papers.nips.cc/paper_files/paper/2017/hash/3f5ee243547dee91fbd053c1c4a845aa-Abstract.html}
}

@article{peyre2019computational,
  author    = {Peyr{\'e}, Gabriel and Cuturi, Marco},
  title     = {Computational Optimal Transport},
  journal   = {Foundations and Trends in Machine Learning},
  volume    = {11},
  number    = {5--6},
  pages     = {355--607},
  year      = {2019},
  doi       = {10.1561/2200000073}
}

@book{efron1993bootstrap,
  author    = {Efron, Bradley and Tibshirani, Robert J.},
  title     = {An Introduction to the Bootstrap},
  publisher = {Chapman and Hall/CRC},
  year      = {1993},
  doi       = {10.1201/9780429246593}
}

@article{holm1979simple,
  author  = {Holm, Sture},
  title   = {A Simple Sequentially Rejective Multiple Test Procedure},
  journal = {Scandinavian Journal of Statistics},
  volume  = {6},
  number  = {2},
  pages   = {65--70},
  year    = {1979}
}

@misc{bhat2024nifty,
  author       = {Bhat, Aparna},
  title        = {Nifty Spot, Futures and Options One-Minute Data from 2017 to 2020},
  year         = {2024},
  publisher    = {Zenodo},
  doi          = {10.5281/zenodo.10899828},
  url          = {https://doi.org/10.5281/zenodo.10899828},
  note         = {CC0 1.0}
}

@article{tzavalis2008recovering,
  author  = {Tzavalis, Elias and Rompolis, Leonidas},
  title   = {Recovering Risk Neutral Densities from Option Prices: A New Approach},
  journal = {Journal of Financial and Quantitative Analysis},
  volume  = {43},
  number  = {4},
  pages   = {1037--1053},
  year    = {2008},
  doi     = {10.1017/S0022109000014435}
}

@incollection{figlewski2008estimating,
  author    = {Figlewski, Stephen},
  title     = {Estimating the Implied Risk Neutral Density for the {U.S.} Market Portfolio},
  booktitle = {Volatility and Time Series Econometrics: Essays in Honor of Robert F. Engle},
  editor    = {Bollerslev, Tim and Russell, Jeffrey R. and Watson, Mark},
  publisher = {Oxford University Press},
  year      = {2008},
  url       = {https://ssrn.com/abstract=1256783}
}

@misc{figlewski2018review,
  author = {Figlewski, Stephen},
  title  = {Risk Neutral Densities: A Review},
  year   = {2018},
  doi    = {10.2139/ssrn.3120028},
  url    = {https://ssrn.com/abstract=3120028}
}

@inproceedings{yang2023mixture,
  title = 	 {Mixture of Normalizing Flows for {E}uropean Option Pricing},
  author =       {Yang, Yongxin and Hospedales, Timothy M.},
  booktitle = 	 {Proceedings of the Thirty-Ninth Conference on Uncertainty in Artificial Intelligence},
  pages = 	 {2390--2399},
  year = 	 {2023},
  editor = 	 {Evans, Robin J. and Shpitser, Ilya},
  volume = 	 {216},
  series = 	 {Proceedings of Machine Learning Research},
  month = 	 {31 Jul--04 Aug},
  publisher =    {PMLR},
  url = 	 {https://proceedings.mlr.press/v216/yang23b.html}
}

@misc{xian2024riskneutral,
  author        = {Xian, Zhonghao and Yan, Xing and Leung, Cheuk Hang and Wu, Qi},
  title         = {Risk-Neutral Generative Networks},
  year          = {2024},
  eprint        = {2405.17770},
  archiveprefix = {arXiv},
  primaryclass  = {q-fin.MF},
  url           = {https://arxiv.org/abs/2405.17770}
}

@article{hernandez2016calibration,
  author  = {Hernandez, Andres},
  title   = {Model Calibration with Neural Networks},
  journal = {SSRN Electronic Journal},
  year    = {2016},
  doi     = {10.2139/ssrn.2812140},
  url     = {https://ssrn.com/abstract=2812140}
}

@article{horvath2021deep,
  author  = {Horvath, Blanka and Muguruza, Aitor and Tomas, Mehdi},
  title   = {Deep Learning Volatility: A Deep Neural Network Perspective on Pricing and Calibration in (Rough) Volatility Models},
  journal = {Quantitative Finance},
  volume  = {21},
  number  = {1},
  pages   = {11--27},
  year    = {2021},
  doi     = {10.1080/14697688.2020.1817974}
}

@article{ruf2020neural,
  author  = {Ruf, Johannes and Wang, Weiguan},
  title   = {Neural Networks for Option Pricing and Hedging: A Literature Review},
  journal = {Journal of Computational Finance},
  year    = {2020},
  doi     = {10.2139/ssrn.3486363},
  url     = {https://ssrn.com/abstract=3486363}
}

@article{shikhman2026diagnosing,
    title={Diagnosing Failure Modes of Neural Operators Across Diverse {PDE} Families},
    author={Lennon Shikhman},
    journal={Transactions on Machine Learning Research},
    issn={2835-8856},
    year={2026},
    url={https://openreview.net/forum?id=0S1LWZHQYn},
    note={}
}

@inproceedings{yang2025hyperiv,
  title = 	 {{H}yper{IV}: Real-time Implied Volatility Smoothing},
  author =       {Yang, Yongxin and Chen, Wenqi and Shu, Chao and Hospedales, Timothy},
  booktitle = 	 {Proceedings of the 42nd International Conference on Machine Learning},
  pages = 	 {70550--70564},
  year = 	 {2025},
  editor = 	 {Singh, Aarti and Fazel, Maryam and Hsu, Daniel and Lacoste-Julien, Simon and Berkenkamp, Felix and Maharaj, Tegan and Wagstaff, Kiri and Zhu, Jerry},
  volume = 	 {267},
  series = 	 {Proceedings of Machine Learning Research},
  month = 	 {13--19 Jul},
  publisher =    {PMLR},
  url = 	 {https://proceedings.mlr.press/v267/yang25d.html}
}

@misc{aboussalah2026finance,
  author        = {Aboussalah, Amine M. and Li, Xuanze and Chi, Cheng and Patel, Raj},
  title         = {Finance-Informed Neural Network: Learning the Geometry of Option Pricing},
  year          = {2026},
  eprint        = {2412.12213},
  archiveprefix = {arXiv},
  primaryclass  = {cs.LG},
  url           = {https://arxiv.org/abs/2412.12213}
}

@misc{zhang2025riskneutral,
  author        = {Zhang, Jian'an},
  title         = {A Risk-Neutral Neural Operator for Arbitrage-Free {SPX--VIX} Term Structures},
  year          = {2025},
  eprint        = {2511.06451},
  archiveprefix = {arXiv},
  primaryclass  = {cs.LG},
  url           = {https://arxiv.org/abs/2511.06451}
}

@misc{conti2026transfer,
  author        = {Conti, Andrea and Morelli, Giacomo},
  title         = {Transfer Learning (Il)liquidity},
  year          = {2026},
  eprint        = {2512.11731},
  archiveprefix = {arXiv},
  primaryclass  = {q-fin.MF},
  url           = {https://arxiv.org/abs/2512.11731}
}

@article{kovachki2023neural,
  author  = {Kovachki, Nikola and Li, Zongyi and Liu, Burigede and Azizzadenesheli, Kamyar and Bhattacharya, Kaushik and Stuart, Andrew and Anandkumar, Anima},
  title   = {Neural Operator: Learning Maps Between Function Spaces with Applications to {PDEs}},
  journal = {Journal of Machine Learning Research},
  volume  = {24},
  number  = {89},
  pages   = {1--97},
  year    = {2023},
  url     = {https://jmlr.org/papers/v24/21-1524.html}
}

@inproceedings{wang2024latent,
  author    = {Wang, Tian and Wang, Chuang},
  title     = {Latent Neural Operator for Solving Forward and Inverse {PDE} Problems},
  booktitle = {Advances in Neural Information Processing Systems},
  volume    = {37},
  pages     = {33085--33107},
  year      = {2024},
  doi       = {10.52202/079017-1042}
}

@inproceedings{raonic2023convolutional,
  author    = {Raonic, Bogdan and Molinaro, Roberto and De Ryck, Tim and Rohner, Tobias and Bartolucci, Francesca and Alaifari, Rima and Mishra, Siddhartha and de B{\'e}zenac, Emmanuel},
  title     = {Convolutional Neural Operators for Robust and Accurate Learning of {PDEs}},
  booktitle = {Advances in Neural Information Processing Systems},
  volume    = {36},
  pages     = {77187--77200},
  year      = {2023}
}

@inproceedings{chen2024dataefficient,
  author    = {Chen, Wuyang and Song, Jialin and Ren, Pu and Subramanian, Shashank and Morozov, Dmitriy and Mahoney, Michael W.},
  title     = {Data-Efficient Operator Learning via Unsupervised Pretraining and In-Context Learning},
  booktitle = {Advances in Neural Information Processing Systems},
  volume    = {37},
  pages     = {6213--6245},
  year      = {2024},
  doi       = {10.52202/079017-0201}
}

@article{arridge2019inverse,
  author  = {Arridge, Simon and Maass, Peter and {\"O}ktem, Ozan and Sch{\"o}nlieb, Carola-Bibiane},
  title   = {Solving Inverse Problems Using Data-Driven Models},
  journal = {Acta Numerica},
  volume  = {28},
  pages   = {1--174},
  year    = {2019},
  doi     = {10.1017/S0962492919000059}
}

@inproceedings{putzky2019invert,
  author    = {Putzky, Patrick and Welling, Max},
  title     = {Invert to Learn to Invert},
  booktitle = {Advances in Neural Information Processing Systems},
  volume    = {32},
  year      = {2019}
}

@article{raissi2019physics,
  author  = {Raissi, M. and Perdikaris, P. and Karniadakis, G. E.},
  title   = {Physics-Informed Neural Networks: A Deep Learning Framework for Solving Forward and Inverse Problems Involving Nonlinear Partial Differential Equations},
  journal = {Journal of Computational Physics},
  volume  = {378},
  pages   = {686--707},
  year    = {2019},
  doi     = {10.1016/j.jcp.2018.10.045}
}

@article{cohen2020detecting,
  author  = {Cohen, Samuel N. and Reisinger, Christoph and Wang, Sheng},
  title   = {Detecting and Repairing Arbitrage in Traded Option Prices},
  journal = {Applied Mathematical Finance},
  volume  = {27},
  number  = {5},
  pages   = {345--373},
  year    = {2020},
  doi     = {10.1080/1350486X.2020.1846573}
}

@inproceedings{yang2020projections,
  author    = {Yang, Shuqi and He, Xingzhe and Zhu, Bo},
  title     = {Learning Physical Constraints with Neural Projections},
  booktitle = {Advances in Neural Information Processing Systems},
  volume    = {33},
  pages     = {5178--5189},
  year      = {2020}
}

@misc{ye2026forecastdojoreplayableenvironmentsbenchmarking,
      title={Forecast-Dojo: Replayable Environments for Benchmarking and Training LLM Forecasting Agents}, 
      author={Liqin Ye and Haorui Wang and Fardin Ahmed and Rongzhi Zhang and Yuan He and Ziyuan Lin and Yanbin Yin and Jing Peng and Michael Galarnyk and Sudheer Chava and Chao Zhang},
      year={2026},
      eprint={2609.28876},
      archivePrefix={arXiv},
      primaryClass={cs.AI},
      url={https://arxiv.org/abs/2609.28876}, 
}
\end{document}